\documentclass[conference,a4paper]{IEEEtran}
\IEEEoverridecommandlockouts

\usepackage[hidelinks]{hyperref}
\usepackage[cmex10]{amsmath}
\usepackage{amssymb,amsfonts}
\usepackage{dblfloatfix}

\usepackage[ruled,vlined]{algorithm2e}
\usepackage{graphicx}
\graphicspath{{Figures/PDF/}{Figures/PNG/}}

\usepackage{booktabs}
\usepackage{siunitx}
\usepackage[numbers,compress]{natbib}
\usepackage{texnames}
\usepackage{bm,bbm}
\usepackage{orcidlink}
\usepackage{multirow}

\begin{document}

\title{Ground-to-Satellite Localization in Unconstrained Image Collections for 3D Scene Reconstruction
\thanks{* The first three authors have equal contribution. 


\vspace{0.5mm}
This work is supported by the Intelligence Advanced Research Projects Activity (IARPA) via Department of Interior/
Interior Business Center (DOI/IBC) contract number
140D0423C0075.}
}

\author{	\IEEEauthorblockN{Angel Daruna$^{*}$, Ben Southall$^{*}$, Niluthpol Chowdhury Mithun$^{*}$, \\ Kshitij Minhas, Nicholas Meegan, Qiao Wang,
Bogdan Matei, Supun Samarasekera, Rakesh Kumar}
\IEEEauthorblockA{\textit{Center for Vision Technologies, SRI International, Princeton, NJ, USA}}
}

\maketitle
\begin{abstract}
	Ground image localization with respect to satellite imagery is a key enabler for metrically-accurate, geo-localized 3D scene reconstruction from unconstrained image collections. Existing cross-view localization methods have strict requirements such as panoramic imagery or known initial locations, limiting their applicability for in-the-wild reconstruction settings. We propose a robust hierarchical cross-view localization framework that leverages geometric constraints from Structure-from-Motion (SfM) models derived from unconstrained ground image collections. Our method generates coarse-to-fine pose hypotheses through a cross-view matching approach and aggregates noisy predictions across SfM model(s) using Kernel Density Estimation to recover consensus alignments while filtering outliers. Experiments demonstrate reliable localization performance from challenging image collections. Empirically we found satellite-referenced alignment enables accurate metric scale estimation, doppelgänger detection, and merging of disjoint SfM reconstructions, resulting in more complete, geo-localized site models than are possible with SfM alone.

\end{abstract}


\section{Introduction}

Vision-based geo-localization is a key enabler for applications such as navigation and 3D reconstruction~\cite{mithun2018learning, daruna2025geosurge, mithun2020rgb2lidar}. Aligning ground-level imagery with satellite imagery provides a global reference frame for large-scale scene modeling. This alignment enables fragmented image collections to be merged into a coherent global model, provides geographic context for resolving pose ambiguities, and supports reconstruction in regions with sparse coverage. However, ground-to-satellite localization remains challenging with unposed image collections.


In real-world reconstruction scenarios, ground images are often collected with heterogeneous cameras, uneven coverage, limited fields of view (FOV), without sequential motion, and no reliable initial poses. The lack of panoramas or sequential motion data severely limits the context needed for cross-view matching, while visually similar but geographically distinct structures (henceforth doppelgängers) often create ambiguous correspondences across regions. 

Existing solutions impose strong assumptions that severely limit their ability to deal with unconstrained data and visual ambiguities. 
Coarse retrieval methods typically rely on panoramic image queries to show reliable performance~\cite{lv2025igarss,zhang2024mt,deuser2023sample4geo}, showing a drastic accuracy drop when restricted to limited-FOV inputs.
Conversely, fine-grained refinement methods require accurate location priors to perform reliably~\cite{wang2023fine,xia2025fg}, making them unsuitable for global localization. Crucially, most approaches treat localization as a single-image task, ignoring the geometric relationships inherent in the underlying scene. Consequently, reconstruction pipelines often revert to relying on strong priors (such as GPS metadata or sequential capture), which are often unavailable in in-the-wild settings.

We address this challenge with a hierarchical framework that explicitly couples localization with reconstruction. Unlike relying on single-image predictions, we exploit SfM submodels built from the ground images to disambiguate noisy matches. Our pipeline first generates coarse-to-fine pose hypotheses for individual images. We then aggregate these predictions within SfM clusters using Kernel Density Estimation (KDE), filtering outliers to recover a robust consensus alignment. This process effectively denoises satellite match-based estimates and provides reliable initialization for global alignment, enabling reliable unconstrained reconstruction.

Experiments show that our method achieves robust geo-localization on unconstrained image collections. Beyond localization accuracy, this geo-alignment enables key reconstruction capabilities that SfM alone cannot reliably provide: recovering metric scale, detecting doppelgängers, and globally aligning disjoint SfM submodels. Together, these benefits lead to a more consistent, scale-accurate, and complete 3D reconstruction of scenes from unconstrained data.



\section{Related Works}

\subsection{Cross-View Ground-to-Satellite Geo-Localization. }
Satellite imagery provides a widely available global reference for geo-localization. Existing methods broadly fall into coarse-grained retrieval and fine-grained pose estimation.

\textbf{Coarse-grained retrieval} methods estimate approximate location and orientation by matching a ground query against a database of satellite tiles. Recent approaches such as Panorama-BEV Co-Retrieval~\cite{ye2024cross}, TransGeo~\cite{zhu2022transgeo}, and Sample4Geo~\cite{deuser2023sample4geo} employ Vision Transformers to learn a shared embedding space (where matching pairs are close) that bridges the large viewpoint gap between ground and satellite imagery.

\textbf{Fine-grained pose estimation} methods aim for precise localization given a coarse location prior. Techniques such as the Correlation-Aware Homography Estimator~\cite{wang2023fine}, DenseFlow~\cite{song2023learning}, and FG2~\cite{xia2025fg} use dense feature correlations to align bird's-eye view (BEV) projections of ground data with satellite orthographic images.

Most existing approaches rely on single-image inputs and often assume panoramic views to reduce ambiguity. In unconstrained settings with limited-FOV cameras, these assumptions lead to noisy and ambiguous estimates. In contrast, we treat cross-view predictions as proposals rather than final outputs, refining them through geometric consensus.





\subsection{Sequence and Multi-Image Localization. } 

To resolve single-image ambiguities, some recent work incorporates temporal or multi-image context. Video-based methods~\cite{zhang2023cross, regmi2021video} aggregate temporal information to smooth trajectories and filter noise. In robotics and AR systems, methods (e.g., \cite{mithun2023cross, di2025uav}) match continuous image streams to satellite data, relying on Visual-Inertial Odometry to enforce precise relative pose constraints between frames.

These approaches depend on temporally continuous data or inertial measurements, which are unavailable in unconstrained image collections. Our method addresses this limitation by leveraging SfM-derived geometric constraints instead of temporal continuity. By robustly integrating SfM submodels with satellite matches via KDE, we enable accurate geo-registration for sparse, unordered, and heterogeneous image datasets.

\section{Method}
\begin{figure}[t]
    \centering
    \vspace{0.1cm}    \includegraphics[width=0.98\linewidth]{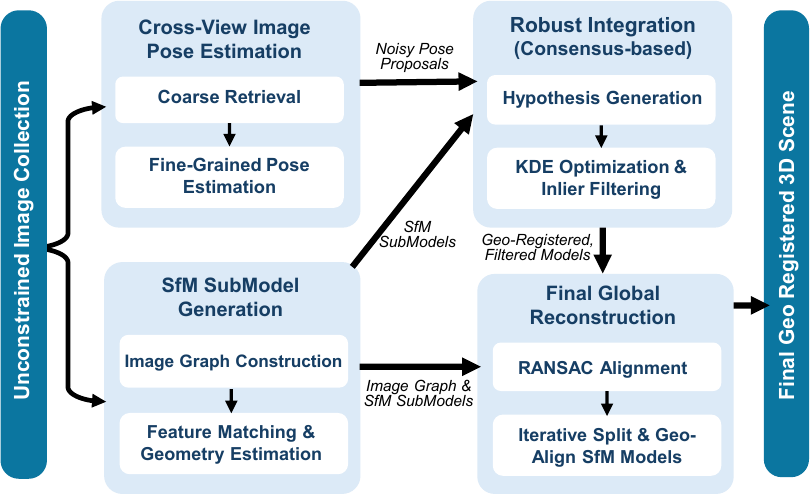}
    \vspace{-0.1cm}
    \caption{A brief illustration of the proposed 3D Reconstruction framework.}
    \label{figure:configuratoin}
    \vspace{-0.3cm}
\end{figure}

We present a robust pipeline (Fig.~\ref{figure:configuratoin}) for 3D reconstruction and geo-localization from image collections, fusing SfM’s relative geometry with noisy absolute cues from ground-to-satellite matching. Next, we discuss the stages of our pipeline.

\subsection{Hierarchical Cross-View Matching}
\label{sec:crossview_match}

We first estimate the absolute pose of each individual image independent of other images in the collection. We develop a coarse-to-fine strategy to generate a distribution of pose hypotheses from fragmented ground-level observations.

\subsubsection{Coarse Retrieval} To identify candidate satellite locations, we follow the approach described in the Panorama-BEV Co-Retrieval Network~\cite{ye2024cross} to train our coarse retrieval model. This approach addresses the drastic geometric domain gap through a dual-branch collaborative retrieval strategy. The network processes the original street view images alongside their transformed BEV projections as query inputs. The street view branch focuses on capturing the global layout to match the broader structural context with satellite images, while the BEV branch emphasizes aligning cross-view shared information with detailed feature matching.
By fusing features from both branches, the model ensures a comprehensive perception of both global layout and local details. For each ground query image $i$, we use this trained model to retrieve the top-$K$ matching satellite tiles $\{L_1, ..., L_K\}$.

\subsubsection{Fine Matching} For each retrieved candidate satellite tile, we estimate the precise 3-DoF pose $(x, y, \theta)$ following method~\cite{wang2023fine}. This method employs a differentiable spherical transform to align the perspective of the ground image with the satellite map, effectively reducing the task to a 2D image alignment problem. To handle challenges such as occlusions, small overlap, and seasonal variations, a robust correlation-aware homography estimator (based on a recurrent convolutional neural network) is utilized to maximize feature correlation between similar regions of the transformed image with the satellite image, ensuring low impact on unobservable areas in alignment. The network directly outputs a homography matrix, from which we derive the 3-DoF pose $P_{sg}^{(i)}$ of the ground camera $i$. Specifically, the metric location is determined by mapping the center point of the transformed ground image to satellite coordinate system, while the orientation is recovered using a reference point above the image's central axis.

\subsection{Structure-from-Motion Submodel Generation}
\label{sec:init_sfm}

Simultaneously to Section~\ref{sec:crossview_match}, we generate local SfM submodels (comprising image locations, camera intrinsic parameters, and sparse 3D points) from the ground image set by organizing the images into pairs with features and correspondences, and supplying these inputs to the COLMAP \cite{schoenberger2016sfm} SfM for relative pose estimation.

We begin by computing global image descriptors (NetVLAD~\cite{arandjelović2016netvladcnnarchitectureweakly}) for all inputs to construct a dense pairwise similarity graph, connecting each image to its $N$ (e.g., $N=40$) most similar neighbors. To mitigate perceptual aliasing, we apply a doppelgänger detection algorithm~\cite{xiangli2025doppelgangers} to identify and prune edges in the image graph between visually similar but geographically distinct images (e.g., repeated architectural elements). While this  sparsifies the graph, highly repetitive scenes often retain incorrect doppelgänger links, which we aim to resolve with our downstream ground-to-satellite matching.
 
We detect local features and match keypoints between remaining connected pairs using the RDD algorithm~\cite{Chen_2025_CVPR}. These correspondences drive the COLMAP geometry estimation pipeline. Depending on the connectivity of the graph, this process yields a set of $M \ge 1$ disjoint submodels $\{ \mathcal{S}_1, ..., \mathcal{S}_M \}$. Ideally, all images form a single coherent model ($M=1$), but fragmentation in the image graph may result in multiple independent clusters. Within each submodel $\mathcal{S}_j$, the relative pose $T_{rel}^{(i)}$ of every image $i$ is known with high precision in a local coordinate system.

\subsection{Robust Alignment via Kernel Density Estimation}


A critical challenge of geo-localizing unconstrained collections is bridging the gap between noisy, single-image matches and a consistent 3D scene model. While the pose proposals from Section~\ref{sec:crossview_match} provide a starting point, limited view context and visual doppelgängers often result in high-variance estimates. We overcome this by leveraging the rigid geometric constraints of the SfM submodels (Section~\ref{sec:init_sfm}).

For a specific submodel, $T_{rel}^{(i)}$ is the relative pose of image $i$ from SfM reconstruction (expressed in the submodel's local coordinate system), and $P_{sg}^{(i)}$ is its absolute predicted noisy pose from from the ground-to-satellite matching stage (Section~\ref{sec:crossview_match}). We can hypothesize the global origin of the submodel, denoted as $H_i$, based on the observation from a single image $i$ as $H_i = P_{sg}^{(i)} \cdot (T_{rel}^{(i)})^{-1}$.

Ideally, all images within the same rigid submodel would produce the same hypothesis $H_i$. However, due to the noisy nature of satellite-ground matches and SfM models, the set of hypotheses $\{H_1, H_2, \dots, H_n\}$ forms a distribution where we expect true matches to cluster together, while alignment errors appear as scattered outliers. To identify the most likely geo-location of the submodel, we treat the hypothesis $H_i$ as samples and estimate the probability density function $\hat{f}(x)$ of the submodel's origin using Kernel Density Estimation. We then solve for the robust global origin, $O_{rst}$, by finding the mode of this density function. $O_{rst}$ represents the geo-coordinate where the most number of images in the submodel agree the submodel origin should be.

This robust estimation of origin allows us to filter outliers. An image $i$ is predicted as an inlier only if its individual hypothesis $H_i$ is within a predefined distance threshold $\delta$ of $O_{rst}$. Hypotheses falling outside this threshold are rejected. The peak density value, $\hat{f}(O_{rst})$, serves as a confidence score for the localization of submodel, where a high value indicates a strong consensus. If confidence score falls below a predefined threshold, the submodel localization is flagged as unreliable.

\subsection{Final Global Reconstruction}\label{subsec:finalglobal}

Given robustly filtered geo-registration results, we can geo-reference the SfM submodels. We transform the satellite matching based location estimates into a local East-North-Up (ENU) coordinate system centered on the mean geo-location. We then align our SfM submodels to these transformed co-ordinates by estimating a Similarity transform (accounting for rotation, translation, and scale) using RANSAC~\cite{RANSAC}.

RANSAC provides a inlier/outlier label for each image location based on alignment error (errors greater than 15 meters are considered outliers). If the RANSAC estimation labels three or more images in a submodel as outliers, we consider it indicating a possible structural inconsistency (e.g., two distinct sub-scenes erroneously merged by SfM). We then split the outlier images into a separate submodel and recursively estimate its geo-location by iterating the process until there are fewer than 3 outlier images left. This process allows us to break apart incorrect SfM sub-models and reassemble the scene into a single geo-referenced output model.

\section{Experiments}

\begin{table}[]
    \centering
    \renewcommand{\arraystretch}{1.2}
    \caption{Brief description of WRIVA datasets referred in experiments}
    \label{tab:wriva_dataset}
    \vspace{-0.1cm}
    \resizebox{0.99\columnwidth}{!}{
        \begin{tabular}{lll}
        \toprule
        \textbf{Datasets} & \textbf{Site Description} \\
        \midrule
        t04\_v01, t05\_v01, t09\_v03 & {Office park building circuit. Johns Hopkins University (JHU) APL, MD}\\
        \midrule
        t04\_v10, t09\_v05 & {Tour of trailer park site, Muscatatuck Training Center (MTC), IN} \\
        \midrule
        t04\_v05, t09\_v04 & {Circuit of lake w/ semi-submerged buildings. Flooded village site, MTC} \\
        \midrule
        t01\_v06 & Shacks, shipping containers, dirt road, long grass. Shantytown site, MTC \\
        \bottomrule
        \end{tabular}
    }
    \vspace{-0.3cm}
\end{table}

We perform our experiments on the WRIVA public datasets ~\cite{wriva_data}. WRIVA datasets have a naming convention t\{TT\}\_v\{VV\}\_s\{SS\}\_r\{RR\}\_\{Name\}, where \{TT\}, \{VV\}, \{SS\}, and \{RR\} specify the theme (type of challenge), vector (location), step (severity of challenge), and revision number respectively. \{Name\} typically describes the theme and site. Descriptions of the sites referred in our experiments are in Table~\ref{tab:wriva_dataset}, with names truncated to theme and vector for brevity.



\text{Implementation Details:} We use ConvNeXt-B as the backbone for coarse retrieval
model, encoding cross-view images, and train the network using the AdamW optimizer over 50 epochs with an initial learning rate of 0.001 and batch size 128. In fine matching, we use EfficientNet-B0 with pretrained weights on Imagenet as the feature extractor for the ground and satellite images. This network is trained using the AdamW optimizer over 100 epochs with an initial leaning rate of 0.00035 and batch size 16. To provide coarse and fine matching with satellite crops of consistent size, we split the large satellite image covering a WRIVA dataset into a grid with a 50\% horizontal and vertical overlap containing cells measuring 70 meters on a side. We derive ortho images from satellite image inputs using the NASA Ames Stereo Processor \cite{ASP, ASP_SHEAN}.

\subsection{Geo-Localization Performance}

We evaluate our hierarchical cross-view matching pipeline on the WRIVA public datasets, splitting the data into 80\% for training and 20\% for testing. We use the same train–test splits across several ablations of our approach, including robust matching that leverages structure-from-motion constraints (+Robust), robust matching combined with failure detection (+Failure Detection), and a baseline variant that excludes both robust matching and failure detection (Ours). Performance is measured using the median error between predicted and ground-truth pose over all test samples, reporting position error in meters and orientation error in degrees. 

\begin{figure}[t]
    \centering
    \includegraphics[width=0.99\linewidth]{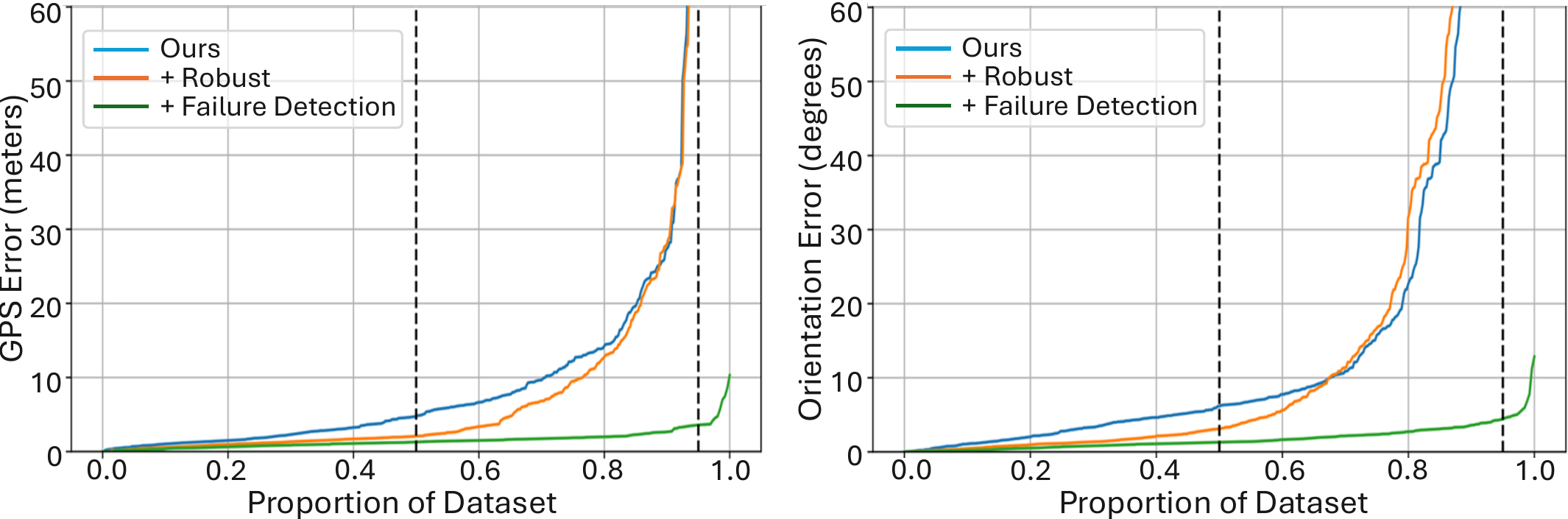}
    \vspace{-0.5cm}
    \caption{Distribution of GPS localization (Left) and orientation/heading (Right) errors over test images within evaluated WRIVA datasets.}
    \label{fig:groundsat_metrics}
    \vspace{-0.5cm}
\end{figure}

The results are summarized in Table~\ref{tbl:groundsat_metrics} and Figure~\ref{fig:groundsat_metrics}: the table presents metrics (i.e., localization and orientation errors) for one sample group from the WRIVA dataset, while the figures show the full distribution of errors for every test image in the WRIVA dataset. Across all datasets, robust matching consistently improves the performance of the cross-view matcher, and failure detection further reduces error by excluding unreliable predictions. Consequently, the full pipeline that incorporates robust matching and failure detection yields the most accurate geo-localization, providing the strongest foundation for improved global reconstruction.

\begin{table}[]
    \centering
    \caption{Errors on WRIVA \textnormal{t09\_v03\_\textbf{Sxx}\_r05\_PTZ\_A01\_SmartCampus} Sets, Sxx $\in \{s00, s01, s02, s03\}$ represents the step number.}
    \small 
    \setlength{\tabcolsep}{3pt}
    \vspace{-0.2cm}
    \resizebox{\linewidth}{!}{%
        \begin{tabular}{lcccccccc}
        \toprule
         & \multicolumn{2}{c}{\textbf{s00}} & \multicolumn{2}{c}{\textbf{s01}} & \multicolumn{2}{c}{\textbf{s02}} & \multicolumn{2}{c}{\textbf{s03}} \\
        \cmidrule(lr){2-3} \cmidrule(lr){4-5} \cmidrule(lr){6-7} \cmidrule(lr){8-9}
        \textbf{Method} & Loc.(m) & Ori.(\si{\degree}) & Loc.(m) & Ori.(\si{\degree}) & Loc.(m) & Ori.(\si{\degree}) & Loc.(m) & Ori.(\si{\degree}) \\ 
        \midrule
        Ours & 11.3 & 11.1 & 10.8 & 11.6 & 9.1 & 8.9 & 7.1 & 7.9 \\
        +Robust & 10.3 & 11.1 & 10.3 & 12.1 & 7.4 & 8.9 & 8.5 & 4.9 \\
        +Fail. Det. & 1.6 & 2.2 & 2.5 & 2.2 & 2.7 & 2.2 & 4.2 & 1.7 \\
        \bottomrule
        \end{tabular}
    }
    \label{tbl:groundsat_metrics}
\end{table}

\begin{table}[t]
\centering
\renewcommand{\arraystretch}{1.05}
\caption{Comparison of Median Localization Errors, and Scale Estimates to show Mutual benefits of satellite/ground and SfM alignment on some randomly selected WRIVA sets.}
\vspace{-0.2cm}
\label{tab:sfm_results}
\setlength{\tabcolsep}{2.5pt} 
\resizebox{0.99\columnwidth}{!}{%
\begin{tabular}{@{\hspace{4pt}}l @{\hspace{10pt}} |ccc |ccc |cc@{}}
\toprule
 & \multicolumn{3}{c|}{\textbf{No. of Images}} & \multicolumn{3}{c|}{\textbf{Median Error (m)}} & \multicolumn{2}{c}{\textbf{Scale Estimate}} \\
\cmidrule(lr){2-4} \cmidrule(lr){5-7} \cmidrule(l){8-9}
 & & All SfM & Valid & Sat/Gnd & SfM & SfM & Aligned & Unaligned \\
\textbf{Set Name} & Input & Localized & Sat/Gnd & (Valid) & (Valid) & (All) & SfM & SfM \\ 
\midrule
t01\_v06\_s03\_r02 & 25  & 20  & 11  & 2.40 & 2.37 & 2.48 & 0.93 & 2.16 \\
t04\_v01\_s00\_r06 & 300 & 295 & 252 & 1.66 & 0.25 & 0.25 & 1.00 & 10.24 \\
t04\_v01\_s01\_r06 & 300 & 283 & 142 & 3.08 & 2.51 & 2.79 & 1.09 & 16.23 \\
t04\_v10\_s01\_r01 & 600 & 599 & 389 & 5.03 & 5.51 & 5.53 & 0.98 & 18.17 \\
t05\_v01\_s00\_r06 & 100 & 100 & 87  & 2.09 & 2.09 & 2.10 & 1.03 & 3.33 \\
t05\_v01\_s01\_r06 & 100 & 97  & 72  & 2.06 & 2.06 & 2.17 & 1.12 & 3.26 \\
t05\_v01\_s02\_r06 & 100 & 87  & 52  & 1.70 & 1.72 & 1.70 & 0.99 & 3.32 \\
t05\_v01\_s03\_r06 & 100 & 87  & 46  & 1.99 & 1.98 & 1.98 & 1.00 & 3.11 \\
t05\_v01\_s04\_r06 & 100 & 83  & 32  & 1.92 & 1.91 & 2.10 & 1.10 & 3.09 \\
t09\_v04\_s00\_r10 & 198 & 189 & 41  & 4.35 & 2.11 & 2.12 & 0.99 & 19.54 \\
t09\_v04\_s01\_r10 & 182 & 175 & 28  & 7.00 & 4.77 & 5.25 & 1.01 & 18.61 \\
t09\_v04\_s02\_r10 & 174 & 163 & 35  & 5.60 & 4.33 & 4.41 & 0.97 & 18.73 \\
t09\_v05\_s00\_r02 & 474 & 471 & 223 & 4.47 & 4.37 & 4.47 & 1.00 & 14.36 \\
t09\_v05\_s01\_r02 & 444 & 440 & 216 & 4.80 & 4.78 & 4.86 & 0.99 & 14.67 \\
t09\_v05\_s03\_r02 & 396 & 393 & 190 & 4.48 & 4.26 & 4.62 & 0.99 & 14.39 \\ 
\bottomrule
\end{tabular}
}
\end{table}

\begin{figure}[t]
    \centering
    \includegraphics[width=0.99\linewidth]{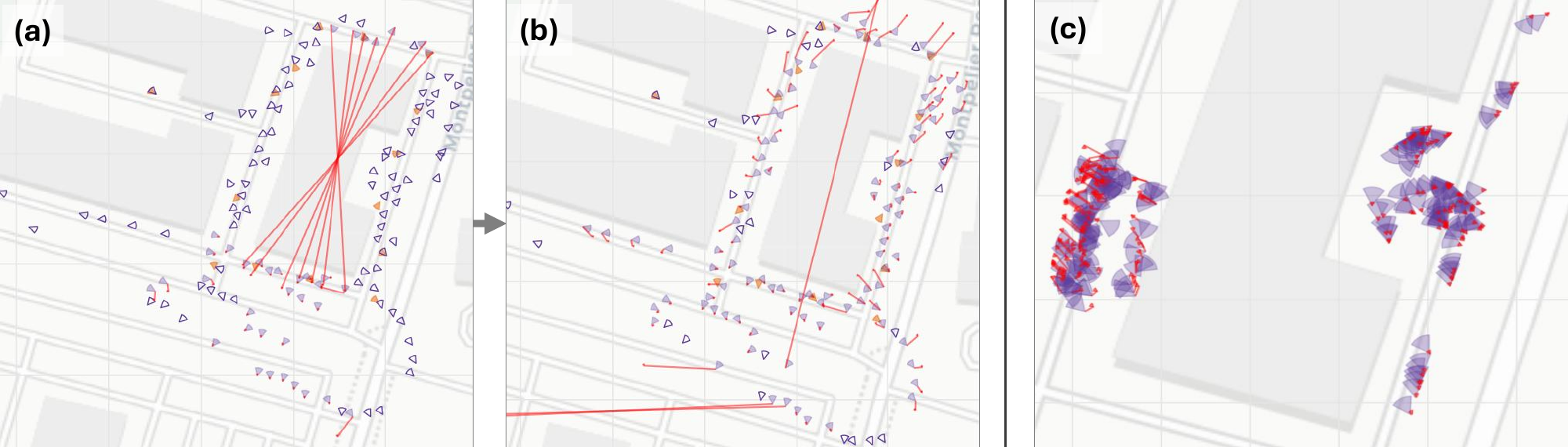}
    \vspace{-0.5cm}
    \caption{SfM Reconstruction on two WRIVA JHU APL sets. (a, b) A large full site set w/o and w/ Sat/Ground alignment. (c) A Two-zone set with alignment.}
    \label{fig:sfm}
    \vspace{-0.3cm}
\end{figure}

\subsection{3D Reconstruction Performance}

We show the mutual benefits of integrating satellite/ground (Sat/Gnd) alignment (adding geo-localization and metric scale to SfM) and SfM models (providing more complete models than Sat/Gnd) by comparing positional alignment error between our Sat/Gnd estimates and our SfM estimates after alignment to Sat/Gnd. The WRIVA datasets include ground truth camera positions in geodetic co-ordinates. We align our SfM model by projecting the Sat/Gnd estimates into a local ENU co-ordinate system (Section \ref{subsec:finalglobal}). For error measurement, we project the ground truth location data into the same co-ordinate system, and then compute the 2D alignment error (since Sat/Gnd does not estimate altitude) between the satellite/ground estimates and the estimates of the aligned SfM. 

Table \ref{tab:sfm_results} shows the median errors of the satellite/ground estimates and the aligned SfM estimates - the (valid) columns show errors \textit{for those images that were successfully aligned in both modalities}. The table also gives the total number of input images for each sequence, and the number of images that were localized in both modalities (as low as 30\% of the total input images). Median errors for the modalities are broadly similar. We observe when satellite/ground fails to localize, the SfM models, with their strong linkage of ground data, can  reconstruct more complete models. This is particularly noticeable in the t05\_v01\_s\{00-04\} test vector, where progressively more images in each step of the vector are corrupted by image artifacts such as blurring, compression, and (over- and under-) saturation. SfM's local feature matcher is able to overcome many of these artifacts, while Sat/Gnd struggles to match the corrupted images. 

The final columns show the scale factor required to bring the SfM model into metric scale, with and without our method - we estimate these factors by fitting a similarity transform between the estimated cameras and their ground truth locations. With our method, model scale is very close to metric (mean 1.0, standard deviation 0.04), while without the method, scales are arbitrary, determined by the SfM process.

Figure \ref{fig:sfm} shows qualitative results of our method on two datasets from the JHU APL site. Input camera locations and orientations are shown as purple wedges; shaded wedges are cameras captured by our final model, outlined wedges are unlocalized cameras. Red lines join the ground truth camera locations to their estimated locations (for plotting, we align to ground truth as for scale estimation). Figure \ref{fig:sfm} (a) shows results  without, and (b) with our method applied to a dataset captured around an office building. Without inputs from satellite/ground, doppelgänger matches across the North and South faces of the building corrupt the model, and many cameras are missed owing to the model fracturing at building corners. With satellite/ground estimates, most doppelgängers are resolved, and many more cameras are included in model. Figure \ref{fig:sfm} (c) shows a result that cannot be obtained without our method - the ground cameras are grouped into two sets with \textit{no visual overlap at all}, and SfM creates two separate models that can \textbf{only} be joined using satellite/ground alignment.

\section{Conclusions}
We introduce a scale-accurate 3D reconstruction framework that integrates SfM with coarse-to-fine satellite matching to resolve geo-localization in unconstrained image collections. By enforcing geometric consistency across SfM clusters, our method aggregates noisy proposals into a robust geographic consensus without relying on temporal data or accurate metadata. Experiments show that this integration significantly improves geo-localization accuracy and enables critical reconstruction capabilities, including metric scale recovery, visual doppelgänger rejection, and merging of disjoint models.

\small
\bibliographystyle{IEEEtranN}
\bibliography{references}
\end{document}